\documentclass{article}
\pdfoutput=1

\usepackage[preprint]{neurips_2021}

\usepackage[utf8]{inputenc} 
\usepackage[T1]{fontenc}    
\usepackage{hyperref}       
\usepackage{url}            
\usepackage{booktabs}       
\usepackage{amsfonts}       
\usepackage{nicefrac}       
\usepackage{microtype}      
\usepackage{xcolor}         
\usepackage{float}
\usepackage{pdfpages}

\usepackage{array}

\title{Towards a Shared Rubric for Dataset Annotation}

\author{%
  Andrew Marc Greene \\
  Adobe\\
  \texttt{agreene@adobe.com} \\
}

\begin{document}

\maketitle

\begin{abstract}
  When arranging for third-party data annotation, it can be hard to
  compare how well the competing providers apply best practices to
  create high-quality datasets. This leads to a ``race to the
  bottom,'' where competition based solely on price makes it hard for
  vendors to charge for high-quality annotation. We propose a
  voluntary rubric which can be used (a)~as a scorecard to compare
  vendors' offerings, (b)~to communicate our expectations of the
  vendors more clearly and consistently than today, (c)~to justify the
  expense of choosing someone other than the lowest bidder, and (d)~to
  encourage annotation providers to improve their practices.
\end{abstract}

\section{Why we need a rubric}

When evaluating data annotation services, the comparison points are
often velocity and price. To compete on those terms, some vendors take
shortcuts that diminish the value of the data, or achieve lower prices
through unethical treatment of the human beings doing the actual
annotation. Discussions of the vendor's processes and their adherence
to best practices sometimes appear on their websites or in
conversation, but it can be difficult to objectively compare
providers. Having a widely used rubric shared by the data community
can formalize these definitions, remind us of the hidden layers in
data sourcing, enable apples-to-apples comparison, and reduce
surprises when the dataset is delivered.

Furthermore, when a project team has selected a vendor who isn't the
lowest bid, they often need to justify their choice to a procurement
team. A scorecard that objectively measures each vendor against a set
of standard crieria should make it easier to explain why spending
additional money is appropriate, in pursuit of essential high-quality
data.

Vendors will also benefit from such a rubric, which can help them more
effectively explain to new clients the value that they provide. It
will reduce the pressure to compete solely on time and money, allowing
a richer ecosystem to thrive with different providers offering
different tradeoffs among price, schedule, and quality.

This rubric is also useful in assessing one's own annotation and
curation practices.

\section{Interpreting the rubric}

We propose 6 major categories (ethical treatment of annotators,
preparation of an ontology and guidelines, assessing annotation
quality, merging of individual annotations, preparing the data for
annotation, and data delivery), several of which have subcategories,
for a total of 15 specific areas.

Each category presents a summary of practices we have encountered with
various vendors, assigned to four levels. (An empty level in a
category means we have not yet encountered a practice that we would
assign to that level.)

\begin{itemize}
\item {\bf Excellent}: This is a best practice and exceeds expectations.
\item {\bf Good}: This is adequate for most cases and is the usual expectation.
\item {\bf Poor}: Below expectations; a warning sign that the provider may deliver poor-quality data.
\item {\bf Unacceptable}: A major deficiency; even one of these usually disqualifies a provider.
\end{itemize}


This rubric, while shared, is not to be mindlessly applied across all
projects. Your team might adjust the ``grade'' for some items to
emphasize what you consider important. We welcome your feedback on
what should be added or judged differently; please send email to
\verb+agreene@adobe.com+
and check
\verb+datacuration.org/rubric+
for the latest version.

Those interested in a detailed explanation of many of these topics
will find Monarch (2021) useful.

\section{The rubric}
\begin{figure}[H]
  \centering
  \includepdf[pages={1}, scale=0.75, offset=0 -50]{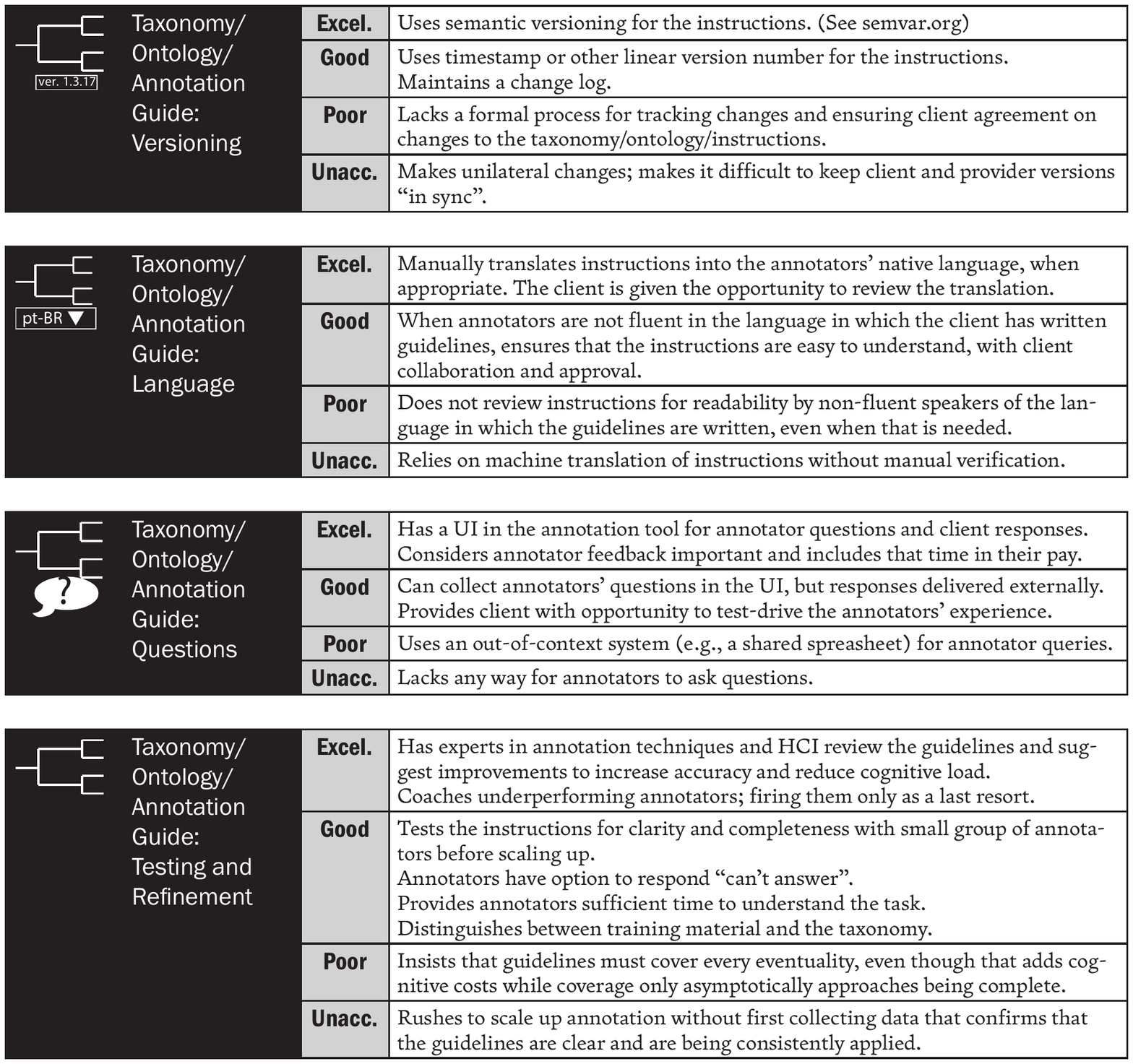}
\end{figure}
\vfill\pagebreak
\begin{figure}[H]
  \centering
  \includepdf[pages={2}, scale=0.75, offset=0 40]{rubric-indd.pdf}
\end{figure}
\pagebreak
\begin{figure}[H]
  \centering
  \includepdf[pages={3}, scale=0.75, offset=0 40]{rubric-indd.pdf}
\end{figure}
\vskip 6.6in




\section*{References}

{\small
  [1] Monarch, Robert.\ (2021) {\it Human-in-the-Loop Machine Learning} Shelter Island: Manning.

}

\vfill\break

\end{document}